\documentclass{article} 
\usepackage[preprint]{colm2025_conference}

\usepackage{microtype}
\usepackage{hyperref}
\usepackage{url}
\usepackage{booktabs}
\usepackage{pifont}
\usepackage{lineno}

\usepackage[table]{xcolor}
\usepackage{xcolor}
\usepackage{tcolorbox}
\usepackage{multirow}
\usepackage{amsmath} 
\usepackage{amssymb}
\usepackage{amsfonts}
\usepackage{makecell}
\usepackage{enumitem}

\definecolor{darkblue}{rgb}{0, 0, 0.5}
\hypersetup{colorlinks=true, citecolor=darkblue, linkcolor=darkblue, urlcolor=darkblue}
\def\xie{\textcolor{black}}
\def\jian{\textcolor{black}}
\newcommand\blfootnote[1]{%
  \begingroup
  \renewcommand\thefootnote{}\footnote{#1}%
  \addtocounter{footnote}{-1}%
  \endgroup
}
\title{RzenEmbed: Towards Comprehensive Multimodal Retrieval}

\author{
Weijian Jian
\quad
Yajun Zhang
\quad
Dawei Liang
\quad
Chunyu Xie
\quad
Yixiao He
\\
\quad\quad\quad\quad\quad\quad\quad\quad\quad\quad\quad\quad
\textbf{Dawei Leng}\textsuperscript{$\dag$}
\quad
\textbf{Yuhui Yin} \\ \\
\quad\quad\quad\quad\quad\quad\quad\quad\quad\quad\quad\quad\quad\quad\quad
360 AI Research
}

\begin{document}

\ifcolmsubmission
\linenumbers
\fi

\maketitle

\begin{abstract}
\blfootnote{\textsuperscript{$\dag$} Corresponding Author, E-mail: lengdawei@360.cn}

The rapid advancement of Multimodal Large Language Models (MLLMs) has extended CLIP-based frameworks to produce powerful, universal embeddings for retrieval tasks. However, existing methods primarily focus on natural images, offering limited support for other crucial visual modalities such as videos and visual documents. To bridge this gap, we introduce RzenEmbed, a unified framework to learn embeddings across a diverse set of modalities, including text, images, videos, and visual documents. We employ a novel two-stage training strategy to learn discriminative representations. The first stage focuses on foundational text and multimodal retrieval. In the second stage, we introduce an improved InfoNCE loss, incorporating two key enhancements. Firstly, a hardness-weighted mechanism guides the model to prioritize challenging samples by assigning them higher weights within each batch. Secondly, we implement an approach to mitigate the impact of false negatives and alleviate data noise. This strategy not only enhances the model's discriminative power but also improves its instruction-following capabilities. We further boost performance with learnable temperature parameter and model souping. RzenEmbed sets a new state-of-the-art on the MMEB benchmark. It not only achieves the best overall score but also outperforms all prior work on the challenging video and visual document retrieval tasks. Our models are available in {\url{https://huggingface.co/qihoo360/RzenEmbed}}.
\end{abstract}

\section{Introduction}
Multimodal retrieval, which aims to find semantically related information across heterogeneous data types like text, images, and video, is a fundamental task in artificial intelligence. 
\xie{Early approaches relied on hand-crafted features and shallow fusion mechanisms, which struggled to capture high-level semantic correspondences. The rise of deep contrastive learning has revolutionized this field, enabling models to learn rich, shared embedding spaces from massive image-text corpora.}

\xie{Landmark models such as CLIP~\citep{CLIP},  
Florence-2~\citep{Florence-2}, and FG-CLIP~\citep{xie2025fg} have demonstrated remarkable zero-shot transfer capabilities by aligning global image and text representations through contrastive objectives. More recently, Multimodal Large Language Models (MLLMs) like LLaVA~\citep{LLaVA} and Qwen2-VL~\citep{Qwen2-VL} have extended these frameworks by leveraging language modeling objectives to produce unified, semantically grounded embeddings. These advances have significantly improved performance on standard retrieval benchmarks, particularly in image-text settings.}

\xie{However, these successes remain largely confined to natural images paired with descriptive text. As we move toward truly universal multimodal systems, there is a growing need to support more complex and structured visual modalities, such as videos with temporal dynamics and visual documents with layout-sensitive semantics. Unfortunately, most existing embedding models are not designed to handle such diversity. When applied to video or document retrieval, they suffer from degraded performance due to misaligned temporal segments, noisy captions, and structural ambiguities. This narrow generalization hinders the development of universal retrieval systems in real-world applications.}

\xie{The Multimodal Embedding Benchmark (MMEB)~\citep{MMEB-v1,MMEB-v2} has emerged to evaluate this broader vision of universal retrieval, requiring strong performance across a heterogeneous suite of tasks. Yet, current methods struggle on the more challenging sub-tasks of MMEB, especially video and visual document retrieval, which presents several technical challenges in their training paradigms.} First, the standard contrastive learning objective can be compromised by the presence of false negatives (semantically similar samples incorrectly treated as negatives) and hard negatives (subtly different samples that the model struggles to distinguish), which impairs the final discriminative ability of the embeddings~\citep{robinson2021contrastivelearninghardnegative}. 
\xie{Second, the temperature parameter in InfoNCE is typically shared or fixed, despite differing optimal scales across tasks (e.g., fine-grained document retrieval may require sharper similarity distributions than coarse video retrieval)~\citep{qiu2023semanticscreatedequalcontrastive}. Third, the design of text prompts significantly influences embedding quality. Untill now, systematic strategies for generating consistent and compact representations remain underexplored~\citep{ju2025generatorembedderharnessinginnate}.}

To address these challenges, we introduce RzenEmbed, a unified framework for learning universal embeddings across text, images, videos, and visual documents.
\xie{Our approach uses a two stage training strategy. The first stage establishes broad cross-modal alignment using diverse multimodal datasets. The second stage refines the model with task-aware improvements, including a hardness-weighted mechanism to reduce the impact of false negatives and emphasize hard negatives, a learnable temperature module for per-task scaling, and a compact embedding  prompt design to ensure discriminative representations. We also apply model souping to improve stability and final performance. On MMEB, RzenEmbed achieves new state-of-the-art results, outperforming all previous methods in overall score and especially in video and visual document retrieval tasks.}

The main contributions of this work are summarized as follows:

\begin{itemize} [leftmargin=*]
    \item We propose RzenEmbed, a unified framework with a two-stage training strategy to learn highly discriminative and universal embeddings for text, images, videos, and visual documents.

    \item We introduce a method to identify and eliminate false negative samples, alongside a hardness-weighted mechanism that enhances the model’s ability to learn from challenging samples.
    
    \item \xie{We integrate a learnable temperature mechanism, a embedding prompt design, and model souping, further improving the model's robustness and performance across diverse modalities.}
   
    \item \xie{We achieve SOTA performance on MMEB, setting new benchmarks in challenging cross-modal retrieval tasks.} 
    
\end{itemize}

\begin{figure}[t!]
	\centering
	\includegraphics[width=0.8\textwidth]{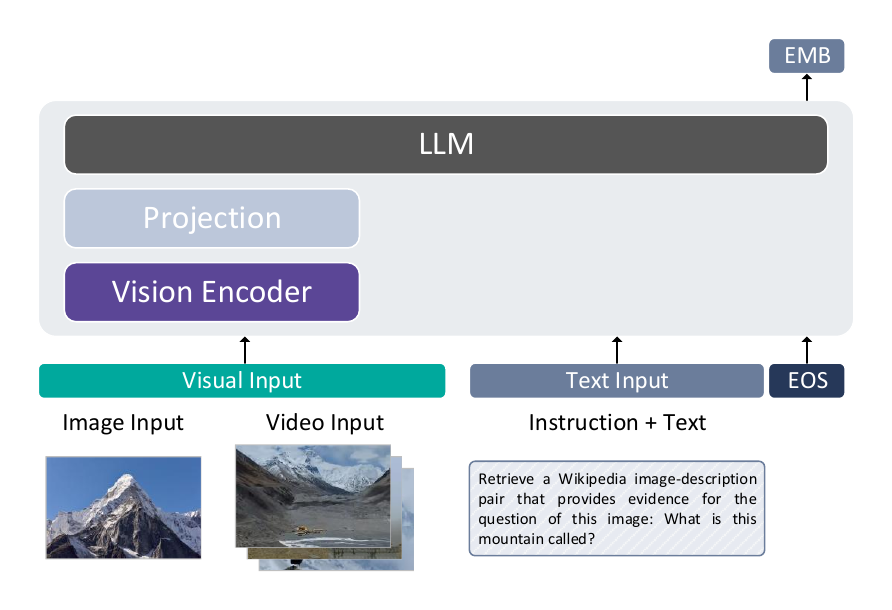} 
	\caption{The architecture of RzenEmbed. The model takes both visual input (images or videos) and text input (instruction + text). Visual data is processed by a vision encoder and projection layer before being fed into the LLM. The LLM jointly encodes both modalities, and the final embedding (EMB) is extracted from the last token's hidden state.}
	\label{fig:one}
\end{figure}

\section{Related Work}
Embedding is an indispensable technique in modern information retrieval, drawing similar items closer while distancing dissimilar ones within a vector space. Traditional methods typically focus on unimodal queries and targets—such as text-to-text, image-to-image, or text-to-image retrieval—and are typically addressed separately using distinct methodologies. The emergence of CLIP~\citep{CLIP} unifies these tasks within a single framework leveraging contrastive learning. Building upon this, \citet{UniIR} introduce the unified instruction-guided retriever UniIR, capable of processing eight distinct tasks with mixed modalities within a single framework. However, UniIR adopts CLIP~\citep{CLIP} and BLIP~\citep{BLIP} as its base models, which exhibit limitations in following complex instructions.

Recently, the advancement of Multimodal Large Language Models (MLLMs) has spurred numerous efforts~\citep{MMEB-v1, Mm-Embed, MegaPairs, GME, mme5, LLaVE, UniME, B3, QQMM-embed} to adapt them for multimodal embedding tasks. \citet{MMEB-v1} introduce MMEB, a benchmark for training and evaluating multimodal embeddings, alongside VLM2Vec, a contrastive framework for converting any MLLM into an embedding model. \citet{Mm-Embed} propose MM-Embed, which enhances text retrieval capabilities through modality-aware hard negative mining and continuous fine-tuning. \citet{MegaPairs} present a data synthesis method, MegaPairs, leveraging public images and vision-language models. Similarly, \citet{GME} introduce a fused-modal data synthesis approach for training a general multimodal embedder. \citet{mme5} further develop a synthesis method covering diverse tasks, languages, and modality combinations, training a multimodal multilingual E5 model. \citet{LLaVE} introduce LLaVE, which improves multimodal embeddings by leveraging the discriminative difficulty of negative pairs. \citet{UniME} propose UniME, a two-stage method: the first stage distills textual knowledge from an LLM teacher to enhance the MLLM's language component, while the second stage employs instruction tuning augmented with hard negatives. \citet{B3} present a novel batch construction technique, B3, constructing a sparse dataset similarity graph and applying community detection to identify clusters of strong negatives. Finally, \citet{QQMM-embed} advocate amplifying gradients for hard negatives within the Info-NCE loss to learn more discriminative multimodal embeddings.

Previous research predominantly focuses on image and text modalities, largely ignoring other visual modalities like video and visual documents. This limitation restricts the practical applicability of these approaches in real-world scenarios. To address these shortcomings, \citet{MMEB-v2} present MMEB-V2, extending the original MMEB benchmark with five novel tasks encompassing video and visual documents. They concurrently propose VLM2Vec-V2, a unified framework designed to learn embeddings across images, videos, and visual documents. Most recently, ByteDance Seed introduce Seed-1.6-Embedding~\citep{bytedance2025}, which employs a three-stage training strategy comprising text continual training, multimodal continual training, and fine-tuning. Seed-1.6-Embedding achieves the highest overall score on the MMEB-V2 benchmark.

\section{Method}
\subsection{Architecture}

Our primary objective is to learn a unified embedding space capable of supporting a diverse range of modalities and tasks. This requires a model backbone that can flexibly encode text, images, and videos, while efficiently processing long-context inputs. Recent advances in Multimodal Large Language Models (MLLMs)~\citep{LLaVA-NeXT,Phi-3,Aquila-VL,PaliGemma,CogVLM2,LLaVA-OneVision,Qwen2-VL,Qwen2.5-VL,InternVL3.5} have demonstrated remarkable performance across various benchmarks and serve as a strong foundation for multimodal embedding systems~\citep{MMEB-v1,ColPali,GME,mme5,LLaVE,MMEB-v2,UniME-V2}.

In light of these requirements, we adopt Qwen2-VL~\citep{Qwen2-VL} as the backbone. As illustrated in Figure 1, the architecture is designed to process both visual and textual data seamlessly. The choice of Qwen2-VL is motivated by its key features that align with our goals: (1) \textbf{Native Dynamic Resolution}, which efficiently handles visual inputs of varying resolutions; (2) \textbf{Multimodal Rotational Position Embeddings (M-RoPE)}, enabling robust modeling of static images and temporal features in videos; and (3) \textbf{Strong Generalization}, particularly for instruction-following tasks. These capabilities make it an ideal choice for scalable and generalizable encoding of heterogeneous multimodal data.

The model accepts two primary types of input: \textbf{visual input} and \textbf{text input}.
\begin{itemize}[leftmargin=*]
	\item \textbf{Visual Input} can be either static images or videos. For video tasks, we represent the video as a sequence of frames sampled at a fixed interval to ensure consistent temporal coverage. 
	\item \textbf{Text Input} is structured as a combination of an Instruction and associated Text. This instruction-based format guides the model to perform specific tasks, such as retrieval or question answering.
\end{itemize}

The Large Language Model (LLM) jointly processes the sequence of projected visual tokens and text tokens. Inspired by established practices in text embedding~\citep{GTE,GME_ref_63}, we extract the embedding (EMB) from the final hidden state of the last token (EOS token) of the entire input sequence. This single vector serves as a comprehensive and unified representation for the given multimodal input.

\subsection{Training}
\subsection{Training Objective}
We train Rzenembed using a contrastive learning framework. Our approach is fundamentally based on the InfoNCE loss~\citep{rusak2025infonceidentifyinggaptheory}, a cornerstone objective in self-supervised and contrastive learning. The core principle of InfoNCE is to train a model that pulls the representation of a query (or "anchor") and its corresponding "positive" sample closer together in the embedding space, while simultaneously pushing it apart from a set of "negative" samples.

Formally, given a query vector $q$, a corresponding positive sample vector $k^+$, and a set of $N$ negative sample vectors $\{k_i^-\}_{i=1}^N$, the InfoNCE loss is defined as:
\begin{equation}
\mathcal{L}_{\text{InfoNCE}} = -\log \frac{\exp(\text{sim}(q, k^+)/\tau)}{\exp(\text{sim}(q, k^+)/\tau) + \sum_{i=1}^N \exp(\text{sim}(q, k_i^-)/\tau)}
\label{eq:infonce}
\end{equation}
Here, $\text{sim}(\cdot)$ denotes a similarity function, typically cosine similarity, and $\tau$ is a temperature hyperparameter that controls the sharpness of the distribution.

Despite its widespread success, the standard InfoNCE loss suffers from two notable limitations in practice:
\begin{itemize}[leftmargin=*]
    \item \textbf{False Negatives:} A training batch may contain samples that are semantically similar to the query but are inadvertently treated as negatives. This penalizes the model for recognizing valid similarities, which can hinder convergence.
    \item \textbf{Dominance of Easy Negatives:} A typical batch is often dominated by easy negatives (samples that are apparently different from the query). This causes the model to allocate most of its learning capacity to trivial distinctions while neglecting the more informative hard negatives, which are crucial for learning a fine-grained representation space.
\end{itemize}
To address these challenges, we introduce two key modifications to the standard InfoNCE framework, as detailed below.

\subsubsection{False Negative Mitigation}
False negatives are instances within a training batch that, despite being sampled as negatives, are semantically similar or even equivalent to the query. For instance, in a batch of text passages, two documents discussing the same topic might be incorrectly contrasted, misleading the model.

To mitigate the impact of false negatives, we adopt a straightforward yet effective filtering strategy. During training, for each query-positive pair $(q, k^+)$, we identify potential false negatives from the set of negative samples $\{k_i^-\}$. A negative sample $k_i^-$ is considered as a false negative if its similarity to the positive sample $k^+$ exceeds a predefined threshold $\delta$:
\begin{equation}
\text{sim}(k_i^-, k^+) > \delta
\end{equation}
These identified false negatives are then excluded from the denominator of the InfoNCE loss calculation (Equation~\ref{eq:infonce}) for that specific query. This simple mechanism prevents the model from being penalized for clustering semantically similar samples together, thereby achieving more stable and meaningful learning.

\subsubsection{Hardness-Weighted Strategy}
Hard negatives are samples that are semantically distinct from the query but lie close to it in the embedding space, making them difficult for the model to distinguish. For example, an image of a Husky might be a hard negative for a query image of a Samoyed, since they are visually similar but belong to different classes. Effectively learning from these challenging examples is critical for developing a robust and discriminative model, as easy negatives offer little learning signal.

To force our model to focus on these informative samples, we incorporate an exponentially hardness-weighted strategy~\citep{robinson2021contrastivelearninghardnegative}. Instead of treating all negatives equally, this method assigns a higher weight to harder negatives in the loss computation. Specifically, we re-weight each negative sample $k_i^-$ based on its similarity to the query $q$. The weight $w_i$ is defined as:
\begin{equation}
w_i = \exp(\alpha \cdot \text{sim}(q, k_i^-))
\end{equation}
where $\alpha > 0$ is a hyperparameter that controls the strength of the weighting. The modified loss function then becomes:
\begin{equation}
\mathcal{L}_{\text{WHNM}} = -\log \frac{\exp(\text{sim}(q, k^+)/\tau)}{\exp(\text{sim}(q, k^+)/\tau) + \sum_{i=1}^N w_i \cdot \exp(\text{sim}(q, k_i^-)/\tau)}
\label{eq:whnm}
\end{equation}
This mechanism ensures that negatives with higher similarity to the query (i.e., harder negatives) receive a larger weight $w_i$, thereby amplifying their contribution to the loss gradient. This effectively directs the model's attention towards learning the fine-grained distinctions necessary to resolve these challenging cases.

By combining false negative elimination and hardness-weighted strategy, our training objective enables Rzenembed to learn a more robust and discriminative embedding space.

\subsection{Recipe}

We train RzenEmbed on a diverse mixture of embedding tasks that span multiple modalities and task types. The training process is divided into two distinct stages: multimodal continual pre-training and fine-tuning. Furthermore, we incorporate a learnable temperature mechanism, a embedding prompt design, and model souping, which further enhance the model's robustness and performance across a wide range of tasks.

\subsubsection{Multimodal Continual Training}
\label{ssec:phase1}

The primary goal of this stage is to equip our model with fundamental embedding capabilities. This involves learning to align representations across text, image, and video modalities into a unified semantic space.

In this stage, we deliberately avoid instruction-based fine-tuning. The sole focus is on developing the model's capacity to generate high-quality and well-aligned embeddings. To this end, we utilize a diverse mixture of training data, categorized into three types:
\begin{itemize}[leftmargin=*]
    \item \textbf{Unimodal Data:} Text-to-Text (T$\rightarrow$T) pairs for improving textual understanding.
    \item \textbf{Cross-modal Data:} Text-to-Image (T$\rightarrow$I) and Text-to-Video Description (T$\rightarrow$VD) pairs for learning cross-modal alignment.
    \item \textbf{Fused-modal Data:} Image-Text-to-Image (IT$\rightarrow$I) pairs, where the model uses a source image and a differential text description to retrieve a target image.
\end{itemize}

For unimodal (T$\rightarrow$T) training, we leverage established datasets such as MS-MARCO~\citep{MARCO}, NQ~\citep{NQ}, HotpotQA~\citep{HotpotQA}, and TriviaQA~\citep{Trivia}. Our cross-modal data is sourced from T$\rightarrow$I pairs in the LAION~\citep{LAION} dataset and T$\rightarrow$VD pairs from ShareGPT4V~\citep{ShareGPT4V}. For fused-modal training (IT$\rightarrow$I), we use the Megapairs dataset~\citep{MegaPairs}, which is specifically designed for this kind of differential image retrieval. To ensure a balanced data distribution during training, we sample from these datasets with the following proportions: 0.3 million T$\rightarrow$T pairs, 0.25 million T$\rightarrow$VD pairs, 2 million T$\rightarrow$I pairs, and 2.5 million fused-modal pairs.

\paragraph{Enhancing Data with Detailed Recaptioning} To improve the model's comprehension of long and detailed text, we enhance our T$\rightarrow$I training data. We use a powerful large multimodal model, CogVLM-19B~\citep{CogVLM2}, to recaption images from the LAION-2B dataset~\citep{LAION}. This strategy fosters a tighter semantic alignment between visual and textual modalities by training on high-quality image-description pairs. This process simultaneously serves as a data-denoising step, yielding representations more robust to the inherent noise of web-crawled captions. Moreover, by replacing generic labels (e.g., \textit{"a cat"}) with fine-grained descriptions (e.g., \textit{"an orange tabby cat basking in the sun"}), our approach enables the model to capture subtle semantic nuances and produce more discriminative embeddings.

Finally, all public datasets undergo a rigorous cleaning process. We employ sophisticated filtering algorithms to remove noise, duplicates, and irrelevant content. We also systematically discard blurry, corrupted, or low-resolution images to ensure the high quality of our training corpus.

\subsubsection{Fine-Tuning}
\label{ssec:phase2}

The objective of this stage is to comprehensively improve the model's ability to handle a wide range of specialized scenarios and complex tasks. We achieve this by introducing a diverse mixture of instruction-formatted data.

We systematically construct a high-quality fine-tuning dataset structured around three key dimensions: \textbf{task type}, \textbf{input modality}, and \textbf{task scenario}. This dataset includes the training set from MMEB-v2~\citep{MMEB-v2}, supplemented by a wide array of public multimodal retrieval and question-answering (QA) datasets. 

Similar to the pre-training stage, all data undergoes a strict cleaning process to ensure high quality. A key aspect of our strategy is that \textbf{each training batch is sampled from single dataset} (except classification dataset). This approach concentrates hard negative samples within each batch, making the contrastive learning objective more effective. The instruction data in this stage is highly diverse and covers a broad spectrum of tasks:
\begin{itemize}[leftmargin=*]
    \item \textbf{For Images:} Tasks include classification, QA (both multiple-choice and open-ended), retrieval, and grounding.
    \item \textbf{For Visual Documents (VisDoc):} The primary task is Visual Document Retrieval.
    \item \textbf{For Videos:} Tasks encompass Video Retrieval, Moment Retrieval, Video Classification, and Video Question Answering.
\end{itemize}
To maintain a balanced task distribution and prevent the model from overfitting to any single task, we cut the number of samples from each individual dataset at 100,000.

\paragraph{Merging Image Classification dataset} The MMEB-v2 training set can be devided into three categories: images, visual documents, and videos. Most image classification datasets have very few categories. For example, HatefulMemes dataset has only 2 categories, VOC2007 dataset has 20 categories, and N24News dataset has 24 categories. During training under contrastive learning, it is necessary to construct an image-text similarity matrix. This results in a large number of false negative samples. Therefore, we merge all image classification datasets into a new dataset. This significantly reduces the number of false negative samples when a batch is sourced from this new dataset.

\paragraph{Enhancing Video Data} We observed that existing video training sets, such as those in MMEB-v2, primarily consist of short videos (under 30 seconds) with high frame-to-frame similarity, making the tasks relatively simple. To address this, we reduce our reliance on this data and supplement our training with a broad collection of public video datasets, processed with the following strategies to increase task difficulty:
\begin{itemize}[leftmargin=*]
    \item \textbf{Segmenting Long Videos:} We divide long videos into multiple short clips, each with a corresponding description. Since these clips originate from the same source video, they serve as natural hard negatives for one another during training.
    \item \textbf{Incorporating Long-Form Videos:} We add long videos (1--3 minutes) paired with holistic descriptions of their overall content. This encourages the model to develop an understanding of long-range temporal dependencies and global context.
\end{itemize}

\subsubsection{Task-Specific Learnable Temperature}

\jian{In contrast to the standard InfoNCE loss, where the temperature \(\tau\) is a manually-tuned hyperparameter, we adopt a learnable temperature, following recent work by \citet{li2023curriculum}. This allows the model to dynamically control the sharpness of the softmax probability distribution during training. The temperature \(\tau > 0\) governs this sharpness: a smaller \(\tau\) creates a sharper distribution, compelling the model to focus on the hardest negative samples, whereas a larger \(\tau\) yields a smoother distribution, encouraging the model to consider all negative samples more uniformly.}

\jian{Our work extends this concept to a large-scale, multi-task setting. Designed for broad multimodal understanding, our training set is organized into seven distinct tasks, including image classification, image question answering, image retrieval, image grounding, document retrieval, video retrieval, and video question answering. Instead of using a single global temperature, we introduce a dedicated, learnable temperature parameter \(\tau_t\) for each task \(t\). This allows the model to learn an optimal, task-specific sharpness, accommodating the varying difficulty and sample distributions across different tasks.}

\jian{To ensure positivity (\(\tau_t > 0\)) and stable optimization, we employ a re-parameterization strategy. For each task-specific temperature \(\tau_t\), we introduce a corresponding learnable scalar \(\theta_t\) and define the temperature as:
\begin{equation}
\tau_t = \exp(\theta_t).
\label{eq:temp}
\end{equation}
This formulation inherently constrains \(\tau_t\) to be positive and allows for the unconstrained optimization of \(\theta_t\) via standard backpropagation, which is updated jointly with other model parameters.
}

\subsubsection{Embedding Prompt}

We leverage Qwen2-VL \citep{Qwen2-VL} in our contrastive learning architecture, which is primarily trained in a generative manner, but this can pose a significant challenge for discriminative representation learning.

To overcome this, we strategically employ a combination of system prompts and representation prompts \citep{ju2025generatorembedderharnessinginnate}, which forces the model to generate representations suitable for discriminative learning.

we use ``Given an image, summarize the provided image in one
word. Given only text, describe the text in one word.'' as the system prompt. And for plain text queries, the representation prompt is ``Represent the given text in one word." , for multimodal queries, the representation prompt is ``Represent the given image in one word."

During model training, the input query is structured as ``$<$system prompt$>$ $<$query$>$ $<$representation prompt$>$".  In inference mode, the query is modified accordingly.

\subsubsection{Model Souping}

We further enhance the model's performance by employing the model souping technique specifically for LoRA adapters~\citep{LoRA,EmbeddingGemma}. Instead of deploying multiple specialized LoRA adapters individually, we first consolidate their learned low-rank weight matrices into a single, generalized adapter through a weighted aggregation or other fusion strategies. This "souped" LoRA adapter then captures the complementary knowledge from the individual adapters. Subsequently, this consolidated LoRA adapter is seamlessly merged with the pre-trained base model, creating a unified and more versatile retrieval model. This approach effectively distills the collective expertise of multiple specialized adapters into a single, efficient entity, significantly reducing computational overhead and memory footprint while preserving or enhancing retrieval performance.

\section{Experiments}

\subsection{Train Data}

Our training methodology is structured in two sequential stages to instill robust textual, cross-modal retrieval, and instruction-following capabilities. In the first stage, we utilized 5 million data entries to develop foundational embedding skills. For text retrieval, this involved incorporating datasets such as MS-MARCO~\citep{MARCO}, NQ~\citep{NQ}, HotpotQA~\citep{HotpotQA}, TriviaQA~\citep{Trivia}, SQuAD~\citep{SQuAD}, FEVER~\citep{FEVER}, and AllNLI for SimCSE~\citep{SimCSE}, totaling approximately 300,000 entries. To enable cross-modal retrieval, we randomly sample 2 million entries from LAION-2B~\citep{LAION}, including both original and CogVLM-19B-generated captions, and supplemented this with 2.5 million randomly sampled MegaPairs dataset entries~\citep{MagicLens} to ensure basic multimodal retrieval proficiency. The second stage primarily focused on training with the MMEB v2 training set, augmented by mmE5-synthetic data and 400,000 video clips sampled from the VideoChat-Flash dataset. This advanced stage aimed to cultivate strong instruction-following retrieval capabilities by exposing the model to diverse multimodal instruction scenarios.

\begin{table*}[!t]
	\centering
    \renewcommand{\arraystretch}{1.2}
	\caption{Results on the MMEB-V1 benchmark~\citep{MMEB-v1}. The results in \textbf{bold} and \underline{underlined} represent the best and second-best performances of different model sizes, respectively. IND: in-distribution, OOD: out-of-distribution. $\textsuperscript{\dag}$: link to the model's homepage.}
	\label{tab:main_exp-v1}
	\resizebox{\textwidth}{!}{
		\begin{tabular}{lccccccccccc}
			\toprule
			\multirow{2}{*}{\textbf{Model}} & \multirow{2}{*}{\textbf{Backbone}} & \multirow{2}{*}{\textbf{Model Size}} & \multicolumn{4}{c}{\textbf{Per Meta-Task Score}} & & \multicolumn{3}{c}{\textbf{Average Score}} \\ 
			\cmidrule(lr){4-7} \cmidrule(lr){9-11}
			& & & \textbf{Classification} & \textbf{VQA}  & \textbf{Retrieval} & \textbf{Grounding} & & \textbf{IND} & \textbf{OOD} & \textbf{Overall} \\ \midrule
			\textbf{\# of datasets} $\rightarrow$ & & & 10 & 10 & 12 & 4 & & 20 & 16 & 36 \\ \midrule
			
			\multicolumn{11}{c}{\textbf{\emph{Encoder-Only Models}}} \\ \midrule
			\rowcolor{gray!10}
			CLIP~\citep{CLIP}   & -  & 0.428B  & 42.8 & 9.1 &  53.0 &  51.8 &   &  37.1  &  38.7 &  37.8 \\
			\rowcolor{white}
			BLIP-2~\citep{BLIP-2}    & - & 3.74B & 27.0  &  4.2 & 33.9  & 47.0 &  &  25.3 &  25.1 & 25.2 \\
			\rowcolor{gray!10}
			SigLIP~\citep{SigLIP}   &  - & 0.203B & 40.3  &  8.4 & 31.6  & 59.5 &  &  32.3 &  38.0 & 34.8 \\
			\rowcolor{white}
			OpenCLIP~\citep{OpenCLIP}   & -  & 0.428B & 47.8  &  10.9 & 52.3  & 53.3 &  &  39.3 &  40.2 & 39.7 \\
			\rowcolor{gray!10}
			UniIR (BLIP\_FF)~\citep{UniIR}   & -  & 0.247B &  42.1 &	 15.0  &	60.1 & 	62.2  &	 & 44.7	&  40.4 & 	42.8 \\
			\rowcolor{white}
			UniIR (CLIP\_SF))~\citep{UniIR}   & -  & 0.428B & 44.3 & 16.2 & 61.8 & 65.3 & & 47.1 & 41.7 & 44.7 \\
			\rowcolor{gray!10}
			Magiclens~\citep{MagicLens}   & -  & 0.428B &  38.8 &  8.3  &  35.4 &  26.0 &  & 31.0 & 23.7  & 27.8  \\
            \midrule
            \multicolumn{11}{c}{\textbf{\emph{Closed-source Models}}} \\ \midrule
            \rowcolor{gray!10}
            Seed-1.6-embedding\href{https://seed1-6-embedding.github.io/}{$\textsuperscript{\dag}$} & Seed1.6-flash  & unknown & 76.1 & 74.0 & 77.9 & 91.3 &  & -  & - & 77.8 \\
			\midrule

			\multicolumn{11}{c}{\textbf{\emph{$\sim$ 2B Models}}} \\ \midrule
			\rowcolor{gray!10}
			VLM2Vec~\citep{MMEB-v1} & Phi-3.5-V  & 4.15B  &54.8  &54.9  &62.3  &79.5  &   & 66.5  &52.0  & 60.1  \\
			VLM2Vec~\citep{MMEB-v1}  & Qwen2-VL  & 2.21B  & 59.0 & 49.4 &  65.4& 73.4 &   & 66.0  & 52.6 & 59.3 \\
			\rowcolor{gray!10}
			VLM2Vec-V2\citep{MMEB-v2}   & Qwen2-VL  & 2.21B  &62.9  &56.3  &69.5  &77.3  &   &  - & - &64.9  \\
			UniME-V2~\citep{UniME-V2}   & Qwen2-VL  & 2.21B  & 62.1 & 56.3 & 68.0 & 72.7 &   &67.4   &58.9  &63.6  \\
			\rowcolor{gray!10}
			GME~\citep{GME}    & Qwen2-VL  & 2.21B  & 54.4 & 29.9 &66.9  &55.5  &   &   - & -  & 51.9 \\
			LLaVE~\citep{LLaVE}   & Aquila-VL  & 1.95B  & 62.1 & 60.2 &65.2  &\underline{84.9}  &   & 69.4  & 59.8 &65.2  \\
			\rowcolor{gray!10}
			B3~\citep{B3}   & Qwen2-VL  &  2.21B &67.0  & 61.2 & \underline{70.9} & 79.9 &   & \underline{72.1}  &\underline{63.1}  & 68.1 \\
			
			UNITE~\citep{UNITE}     & Qwen2-VL  &  2.21B &63.2  &55.9  & 65.4 &75.6  &   &65.8   & 60.1 & 63.3 \\
			\rowcolor{gray!10}
			ColPali-v1.3 \citep{ColPali}    & PaliGemma &  2.92B &40.3  & 11.5 &48.1  & 40.3 &   &-   & - &34.9  \\
			CAFe~\citep{CAFe} & LLaVA-OV & 0.894B & 59.1 & 49.1 & 61.0 & 83.0 &   & 64.3  & 53.7 & 59.6 \\
			Ops-MM-embedding-v1\href{https://huggingface.co/OpenSearch-AI/Ops-MM-embedding-v1-2B}{$\textsuperscript{\dag}$}   & Qwen2-VL &  2.21B & \underline{68.1} & \underline{65.1} & 69.2 & 80.9 &   &  - &-  &\underline{69.0}  \\
			\rowcolor{blue!5}
			\textbf{RzenEmbed (ours)}   & Qwen2-VL  & 2.21B  &\textbf{68.5}  & \textbf{66.3} & \textbf{74.5} & \textbf{90.3} &  & \textbf{76.1}   & \textbf{67.4} & \textbf{72.3} \\
			
			\midrule	
			\multicolumn{11}{c}{\textbf{\emph{$\sim$ 7B Models}}} \\ \midrule
			\rowcolor{gray!10}
			VLM2Vec~\citep{MMEB-v1}   & LLaVA-1.6  & 7.57B  & 61.2 &49.9 & 67.4 &86.1  &   & 67.5  & 57.1 & 62.9 \\
			VLM2Vec~\citep{MMEB-v1}   & Qwen2-VL  & 8.29B  & 62.6 & 57.8 & 69.9 & 81.7 &   & 65.2  & 56.3 & 65.8 \\
			
			\rowcolor{gray!10}
			UniME-V2~\citep{UniME-V2}    & LLaVA-OV  & 8.03B  & 65.3 & 67.6 & 72.9 & 90.2 &   & 74.8  & 66.7 & 71.2 \\
			UniME-V2~\citep{UniME-V2}    & Qwen2-VL  & 8.29B  & 64.0 &60.1  & 73.1 & 82.8 &   & 72.0  &63.0  &68.0  \\	
			\rowcolor{gray!10}
			GME~\citep{GME}   & Qwen2-VL  & 8.29B  & 57.7 & 34.7 & 71.2 & 59.3 &   &  - & - &56.0  \\
			LLaVE~\citep{LLaVE}    & LLaVA-OV  & 8.03B  & 65.7 & 65.4 &  70.9&  \underline{91.9} &  &  75.0 & 64.4 & 70.3 \\
			
			\rowcolor{gray!10}
			B3~\citep{B3}   & Qwen2-VL  &  8.29B & \underline{70.0} &66.5  & \underline{74.1} & 84.6 &   &\underline{75.9}   &\underline{67.1}  & 72.0 \\
			UNITE~\citep{UNITE}    & Qwen2-VL  & 8.29B  &68.3  & 65.1 & 71.6 & 84.8 &   & 73.6  & 66.3 &70.3  \\
			
			\rowcolor{gray!10}
			QQMM-embed~\citep{QQMM-embed}  & LLaVA-OV  &  8.297B & 66.8 & 66.8 & 70.5 &  90.4 &   &  74.7 & 65.6 & 70.7 \\
			CAFe~\citep{CAFe} & LLaVA-OV & 8.03B & 65.2 &65.6 & 70.0 & 91.2 &   & 75.8   & 62.4 &69.8  \\
			
			\rowcolor{gray!10}  
			Ops-MM-embedding-v1\href{https://huggingface.co/OpenSearch-AI/Ops-MM-embedding-v1-7B}{$\textsuperscript{\dag}$}     & Qwen2-VL &  8.29B & 69.7 &\underline{69.6}  & 73.1 & 87.2 &   &  - &-  &\underline{72.7}  \\
			LamRA\href{https://huggingface.co/code-kunkun/LamRA-Ret}{$\textsuperscript{\dag}$}  & Qwen2-VL  & 8.29B  & 59.2 & 26.5 &  70.0& 62.7 &   & -  & - & 54.1 \\
			
			\rowcolor{gray!10}
			LamRA\href{https://huggingface.co/code-kunkun/LamRA-Ret-Qwen2.5VL-7b}{$\textsuperscript{\dag}$} & Qwen2.5-VL  & 8.29B  & 51.7 & 34.1 &66.9 & 56.7 &   &  -  & -  & 52.4 \\
			\rowcolor{blue!5}
			\textbf{RzenEmbed (ours)}  & Qwen2-VL  & 8.29B   & \textbf{70.6} & \textbf{71.7} & \textbf{78.5} & \textbf{92.1} &   & \textbf{78.5}  & \textbf{72.7} & \textbf{75.9} \\
			
			\bottomrule
		\end{tabular}
	}
\end{table*}
\subsection{Training Configuration}
We fine-tune the model for a single epoch using the AdamW optimizer. For parameter-efficient tuning, we apply Low-Rank Adaptation (LoRA) to all linear layers of both the vision encoder and the Large Language Model (LLM), with a uniform rank of 64. 
The learning rate is initialized to 2e-4 and decayed following a cosine schedule. We use a global batch size of 768 and a weight decay of 5e-2. For our proposed training strategies, we set the reweighting factor $\alpha=9$ for the hardness-weighted strategy and the threshold $\delta=0.95$ for the false negative elimination strategy. The maximum number of input tokens for both images and videos is 1280. To enhance memory efficiency, we employ bfloat16 (bf16) mixed-precision training and enable gradient checkpointing. 
All experiments are conducted on 16~NVIDIA A800 (80GB) GPUs.

\subsection{Main Results}
We evaluate RzenEmbed on a comprehensive suite of benchmarks that span diverse task types and modalities. Specifically, our evaluation results on the MMEB-V1~\citep{MMEB-v1} and MMEB-V2~\citep{MMEB-v2} benchmarks are reported in Table~\ref{tab:main_exp-v1} and Table~\ref{tab:main_exp-v2}, respectively.

\paragraph{Results on MMEB-V1} We report RzenEmbed's performance on MMEB-V1 in Table~\ref{tab:main_exp-v1}, alongside a comparison with recent related works. The results reveal that RzenEmbed achieves the best performance in both 2B and 7B model scales. Moreover, RzenEmbed attains optimal performance across Per Meta-Task and in both OOD and IND task divisions, which illustrates RzenEmbed's exceptional adaptability to numerous tasks and its generalization power over data from different domains.

\paragraph{Results on MMEB-V2} The results  in Table~\ref{tab:main_exp-v2} demonstrate that RzenEmbed achieves excellent performance across tasks involving different input modalities, including images, videos, and visual documents on MMEB-V2. Overall, compared to the next best models of same scale, RzenEmbed 2B and 7B models show improvements of 3.4\% and 4.0\%, respectively. Notably, RzenEmbed's 7B model outperforms the closed-sourced Seed-1.6-embedding on both the Video and VisDoc subtasks, as well as achieving a higher overall score on MMEB-V2. Analyzing individual tasks, RzenEmbed consistently achieves top performance in 9 and 11 tasks for the 2B and 7B versions, respectively, with a slight underperformance on a few video meta tasks. This further highlights RzenEmbed's comprehensive multimodal representation capabilities.
\begin{table}[!t]
	\centering
	\renewcommand{\arraystretch}{1.2}
	\huge
	\caption{Results on the MMEB-V2 benchmark~\citep{MMEB-v1}. The results in \textbf{bold} and \underline{underlined} represent the best and second-best performances of different model sizes, respectively. CLS: classification, QA: question answering, RET: retrieval, GD: grounding, MRET: moment retrieval, VDR: ViDoRe, VR: VisRAG, OOD: out-of-distribution. $\textsuperscript{\dag}$: link to the model's homepage.}
\label{tab:main_exp-v2}
\resizebox{\textwidth}{!}{
	\begin{tabular}{l cc ccccc ccccc ccccc c}
		\toprule
		\multirow{2}{*}{\textbf{Model}} 
		& \multirow{2}{*}{\textbf{Backbone}} & \multirow{2}{*}{\textbf{Model Size}}
		& \multicolumn{5}{c}{\textbf{Image}} 
		& \multicolumn{5}{c}{\textbf{Video}} 
		& \multicolumn{5}{c}{\textbf{VisDoc}}
		& \multirow{2}{*}{\textbf{All}}\\
		\cmidrule(lr){4-8} \cmidrule(lr){9-13} \cmidrule(lr){14-18}
		& & & \textbf{CLS} & \textbf{QA} & \textbf{RET} & \textbf{GD} & \textbf{Overall} 
		& \textbf{CLS} & \textbf{QA} & \textbf{RET} & \textbf{MRET} & \textbf{Overall} 
		& \textbf{VDRv1} & \textbf{VDRv2} & \textbf{VR} & \textbf{OOD} & \textbf{Overall} \\
		\midrule
		\textbf{\# of Datasets} $\rightarrow$ & &
		& 10 & 10 & 12 & 4 & 36 
		& 5 & 5 & 5 & 3 & 18 
		& 10 & 4 & 6 & 4 & 24
		& 78 
		\\

        \midrule
        \multicolumn{18}{c}{\textbf{\emph{Closed-source Models}}} \\ \midrule
        \rowcolor{gray!10}
        Seed-1.6-embedding\href{https://seed1-6-embedding.github.io/}{$\textsuperscript{\dag}$} & Seed1.6-flash  & unknown &
					76.1 & 74.0 &  77.9& 91.3 & 77.8 & 
					55.0 & 60.9 & 51.3 & 53.5 & 55.3 & 
					85.5 & 56.6 & 84.7 & 43.1 & 73.4 & 71.3\\
        \midrule	
        
		\multicolumn{18}{c}{\textbf{\emph{$\sim$ 2B Models}}} \\
		\midrule
		
		\rowcolor{gray!10}
		VLM2Vec~\citep{MMEB-v1} & Qwen2-VL  & 2.21B  &
		58.7 & 49.3 & 65.0 & 72.9 & 59.7 & 
		33.4 & 30.5 & 20.6 & 33.0 & 29.0 & 
		49.8 & 13.5 & 51.8 & 33.5 & 41.6 & 
		47.0    \\
		VLM2Vec-V2~\citep{MMEB-v2} & Qwen2-VL  & 2.21B &
		62.9 & 56.3 & \underline{69.5} & 77.3 & 64.9 &
		39.3 & 34.3 & 28.8 & 38.5 & 34.9 & 
		75.5 & 44.9 & 79.4 & 39.4 & 65.4 & 
		58.0 \\
		
		\rowcolor{gray!10}
		GME~\citep{GME}   & Qwen2-VL  & 2.21B  &
		54.4 & 29.9 & 66.9 & 55.5 & 51.9 & 
		34.9 & 42.0 & 25.6 & 32.4 & 33.9 & 
		86.1 & 54.0 & 82.5 & 43.1 & 72.7 & 
		54.1      \\

		ColPali-v1.3~\citep{ColPali}    & PaliGemma &  2.92B   &
		40.3 & 11.5 & 48.1 & 40.3 & 34.9 & 
		26.7 & 37.8 & 21.6 & 25.5 & 28.2 & 
		83.6 & 52.0 & 81.1 & 43.1 & 71.0 & 
		44.4
		\\
		\rowcolor{gray!10}
		CAFe~\citep{CAFe} & LLaVA-OV & 0.894B &
		56.4& 45.3 & 57.6 & 72.0 & 55.4 & 
		33.9 &41.7  & 29.7 & \textbf{39.7} & 35.9 & 
		56.9& 32.6 & 68.6 & 30.7 & 51.4 & 
		49.7 \\
		
		Ops-MM-embedding-v1\href{https://huggingface.co/OpenSearch-AI/Ops-MM-embedding-v1-2B}{$\textsuperscript{\dag}$}    & Qwen2-VL  & 2.21B    &
		\underline{68.1} & \underline{65.1} & 69.2 & \underline{80.9} & \underline{69.0} &
		\textbf{53.6} & \textbf{55.7} & \underline{41.8} & 33.7 & \textbf{47.6} &
		\underline{87.0} & \textbf{57.6} & \underline{85.4} & \underline{43.3} & \underline{74.4} &
		\underline{63.4}
		
		\\

		\rowcolor{blue!5}
		\textbf{RzenEmbed (ours)}   & Qwen2-VL  & 2.21B  & 
		\textbf{68.5}& \textbf{66.3} & \textbf{74.5} & \textbf{90.3} & \textbf{72.3} & 
		\underline{50.4}& \underline{49.7} & \textbf{46.6} & \underline{38.9} & \underline{47.3} & 
		\textbf{87.1}&  \underline{55.1}&   \textbf{87.2}&  \textbf{43.4}&\textbf{74.5} &\textbf{67.2}
		
		\\

		\midrule
		
		\multicolumn{18}{c}{\textbf{\emph{$\sim$ 7B Models}}} \\
		\midrule
		\rowcolor{gray!10}
		VLM2Vec~\citep{MMEB-v1}  & Qwen2-VL  & 8.29B &
		62.7 & 56.9 & 69.4 & 82.2 & 65.5 & 
		39.1 & 30.0 & 29.0 & 40.6 & 34.0 & 
		56.9 & 9.4 & 59.1 & 38.1 & 46.4 & 
		52.3 \\
		
		GME~\citep{GME}   & Qwen2-VL  & 8.29B  &
		57.7 & 34.7 & 71.2 & 59.3 & 56.0 & 
		37.4 & 50.4 & 28.4 & 38.2 & 38.6 & 
		\underline{89.4} & 55.6 & \underline{85.0} & \textbf{44.4} & \underline{75.2} & 
		57.8      \\
		
		\rowcolor{gray!10}
		CAFe~\citep{CAFe} & LLaVA-OV & 8.03B &
		63.6 & 61.7 & 69.1 & \underline{87.6} & 67.6 & 
		35.8& 58.7 & 34.4 & 39.5 & 42.4 & 
		70.7& 49.6 & 79.5 &38.1  & 63.9 & 
		60.6 \\
		
		Ops-MM-embedding-v1\href{https://huggingface.co/OpenSearch-AI/Ops-MM-embedding-v1-7B}{$\textsuperscript{\dag}$}    & Qwen2-VL& 8.29B    &
		\underline{69.7} & \underline{69.6} & \underline{73.1} & 87.2 & \underline{72.7} &
		\textbf{59.7} & \underline{62.2} & \underline{45.7}  &  \underline{43.2}  & \underline{53.8} &
		80.1 & \underline{59.6}  &79.3 & 43.3 & 70.3 &
		\underline{67.6}
		\\
		
		\rowcolor{gray!10}
		LamRA\href{https://huggingface.co/code-kunkun/LamRA-Ret}{$\textsuperscript{\dag}$} & Qwen2-VL  & 8.29B &
		59.2 & 26.5 & 70.0 & 62.7 & 54.1 & 
		39.3 & 42.6 & 24.3 & 34.6 & 35.2 & 
		22.0 & 11.5 & 37.4 & 21.0 & 23.9 & 
		40.4  \\
		LamRA\href{https://huggingface.co/code-kunkun/LamRA-Ret-Qwen2.5VL-7b}{$\textsuperscript{\dag}$}  & Qwen2.5-VL & 8.29B &
		51.7 & 34.1 & 66.9 & 56.7 & 52.4 & 
		32.9 & 42.6 & 23.2 & 37.6 & 33.7 & 
		56.3 & 33.3 & 58.2 & 40.1 & 50.2 & 
		47.4 \\

		\rowcolor{blue!5}
		\textbf{RzenEmbed (ours)}   & Qwen2-VL  & 8.29B  &
		\textbf{70.6}& \textbf{71.7} & \textbf{78.5} & \textbf{92.1} & \textbf{75.9} & 
		\underline{58.8}& \textbf{63.5} & \textbf{51.0} & \textbf{45.5} & \textbf{55.7} & 
        \textbf{89.7} & \textbf{60.7} & \textbf{88.7} & \textbf{44.4} & \textbf{77.1}&
		\textbf{71.6}\\
		\bottomrule
	\end{tabular}
}
\end{table}

\begin{table}[!t]
	\centering
    \small
	\renewcommand{\arraystretch}{1.2}
	\caption{Ablations of strategies.  The results in \textbf{bold} represent the best performances of different strategies. }
	\label{tab:ablation_exp-mixtures-2}
    
    \resizebox{\textwidth}{!}{
	\begin{tabular}{l cccccccc}
		\toprule
		\textbf{Strategies} & \textbf{  \makecell[c]{Merging \\ classification dataset}} & \textbf{ \makecell[c]{ Learnable \\ temperature }} & \textbf{\makecell[c]{ System \\ prompt }} & \textbf{\makecell[c]{ Dataset \\ Resample }} & \textbf{Overall} & \textbf{Image}& \textbf{Video}& \textbf{Visdoc} \\
		\midrule
		Baseline & \ding{55} & \ding{55} & \ding{55} & \ding{55} & 65.7 & 71.4 & 43.5 & 73.8 \\
	    Exp1 & \ding{51} & \ding{55} & \ding{55} & \ding{55} & 66.3 & 71.0 & 45.3 & 75.0 \\
		Exp2 & \ding{55} & \ding{51} & \ding{55} & \ding{55} & 66.4 & 71.5 & 44.0 & \textbf{75.3} \\
            Exp3 & \ding{55} & \ding{55} & \ding{51} & \ding{55} & 66.4 & 71.3 & 45.0 & 75.0 \\
            Exp4 & \ding{51} & \ding{51} & \ding{51} & \ding{55} & 66.7 & 71.6 & 45.8 & 75.0 \\
            Exp5 & \ding{51} & \ding{51} & \ding{51} & \ding{51} & \textbf{67.2} & \textbf{72.3} & \textbf{47.3} & 74.5 \\
		\midrule

	\end{tabular}
    }
\end{table}

\subsection{Ablations of Strategies}
In this section, we conduct a series of ablation studies to validate the effectiveness of each proposed strategy. The results are summarized in Table~\ref{tab:ablation_exp-mixtures-2}. Our baseline model is trained with a standard InfoNCE loss and achieves an overall score of 65.7.

\paragraph{Effect of Merging Classification Datasets} We hypothesize that the limited label space of individual image classification datasets may restrict the model's semantic understanding. To address this, we merge multiple classification datasets into a single, larger one with a richer label set. As shown in Table~\ref{tab:ablation_exp-mixtures-2} (Exp1), this strategy improves the overall performance to 66.3. Notably, it enhances performance on video and VisDoc retrieval, suggesting that a broader semantic foundation for images benefits cross-modal learning.

\begin{table}[!t]
	\centering
    \small
	\renewcommand{\arraystretch}{1.2}
	\caption{Results using different training mixtures. The best results are shown in \textbf{bold}.}
	\label{tab:ablation_exp-mixtures-1}
	\begin{tabular}{l cccc}
		\toprule
		\textbf{Pooling} & \textbf{Overall} & \textbf{Image-Overall} & \textbf{Video-Overall} & \textbf{Visdoc-Overall}
		\\
		\midrule
		Mix 1 &71.11 &75.78 &54.16 &76.83 \\
		Mix 2 &71.16 &75.64 &54.59 &76.86 \\
		Mix 3 &71.18 &75.43 &55.09 & 76.88\\
		\midrule
		\rowcolor{blue!5}
		\textbf{Souped} & \textbf{71.61} & \textbf{75.92} & \textbf{55.73} & \textbf{77.06} \\
		
		\bottomrule
	\end{tabular}
\end{table}

\paragraph{Effect of Learnable Temperature} Given the significant heterogeneity in the data distributions and task formats of our training datasets, a single, fixed temperature for the InfoNCE loss is suboptimal. We introduce task-specific learnable temperatures, grouping datasets into seven distinct tasks (e.g., image classification, VQA, retrieval). This allows the model to dynamically balance the penalties for negative samples across different tasks. Table~\ref{tab:ablation_exp-mixtures-2} (Exp2) shows this mechanism lifts the overall score to 66.4 and achieves the best VisDoc performance (75.3), confirming the benefits of adaptive temperature scaling.

\paragraph{Effect of System Prompt} Our model, Rzenembed, utilizes Qwen2-VL, a backbone pre-trained on generative tasks. To better adapt it for discriminative retrieval tasks, we employ an instruction-tuning approach inspired by prior work \cite{ju2025generatorembedderharnessinginnate}. Specifically, we prepend a system prompt, "summarize the user's intent in one word," to the query. This simple instruction guides the model to produce more discriminative embeddings. As seen in Table~\ref{tab:ablation_exp-mixtures-2} (Exp3), this strategy alone improves the performance to 66.4, demonstrating its effectiveness in bridging the gap between generative pre-training and discriminative fine-tuning.

\paragraph{Effect of Dataset Resampling} During training, we observed that the loss on video datasets converged much faster than on image datasets, indicating an imbalance in learning dynamics. To mitigate this, we implement a dataset resampling strategy to increase the sampling ratio of image-related data. This rebalancing allows the model to learn more effectively from the slower-converging tasks. As shown in the final experiment (Exp5), when all strategies are combined, resampling further boosts the performance to our best result of 67.2. This highlights the importance of balancing the training data exposure based on task-specific convergence rates.

\subsection{Results of Model Souping}
We also explore the effectiveness of model souping for LoRA adapters, which involves merging multiple specialized adapters into a single, generalized one. This consolidated adapter captures complementary knowledge, leading to improved performance. As shown in Table~\ref{tab:ablation_exp-mixtures-1}, this strategy yields the best overall score of 71.61.

\section{Conclusion}
In this paper, we present RzenEmbed, a novel unified framework that significantly advances multimodal embedding learning. By introducing a sophisticated two-stage training strategy, including a hardness-weighted InfoNCE loss with false negative mitigation, RzenEmbed effectively learns discriminative and universal representations across text, images, videos, and visual documents. Extensive experimental evaluations confirm RzenEmbed's superior performance. It achieves state-of-the-art results on the MMEB leaderboard, setting new records in visual document retrieval, video retrieval, and overall score. As a compact yet highly effective model, RzenEmbed provides a powerful solution to the growing need for advanced multimodal retrieval in applications such as AI agents, multimodal search and recommendation, and Retrieval-Augmented Generation.

\bibliography{references_main}

\begin{thebibliography}{52}
\providecommand{\natexlab}[1]{#1}
\providecommand{\url}[1]{\texttt{#1}}
\expandafter\ifx\csname urlstyle\endcsname\relax
  \providecommand{\doi}[1]{doi: #1}\else
  \providecommand{\doi}{doi: \begingroup \urlstyle{rm}\Url}\fi

\bibitem[Abdin et~al.(2024)Abdin, Jacobs, Awan, Aneja, Awadallah, Awadalla, Bach, Bahree, Bakhtiari, Behl, Benhaim, Bilenko, Bjorck, Bubeck, Cai, Mendes, Chen, Chaudhary, Chopra, Giorno, de~Rosa, Dixon, Eldan, Iter, Garg, Goswami, Gunasekar, Haider, Hao, Hewett, Huynh, Javaheripi, Jin, Kauffmann, Karampatziakis, Kim, Khademi, Kurilenko, Lee, Lee, Li, Liang, Liu, Lin, Lin, Madan, Mitra, Modi, Nguyen, Norick, Patra, Perez{-}Becker, Portet, Pryzant, Qin, Radmilac, Rosset, Roy, Ruwase, Saarikivi, Saied, Salim, Santacroce, Shah, Shang, Sharma, Song, Tanaka, Wang, Ward, Wang, Witte, Wyatt, Xu, Xu, Yadav, Yang, Yang, Yu, Zhang, Zhang, Zhang, Zhang, Zhang, Zhang, Zhang, and Zhou]{Phi-3}
Marah~I Abdin, Sam~Ade Jacobs, Ammar~Ahmad Awan, Jyoti Aneja, Ahmed Awadallah, Hany Awadalla, Nguyen Bach, Amit Bahree, Arash Bakhtiari, Harkirat~S. Behl, Alon Benhaim, Misha Bilenko, Johan Bjorck, S{\'{e}}bastien Bubeck, Martin Cai, Caio C{\'{e}}sar~Teodoro Mendes, Weizhu Chen, Vishrav Chaudhary, Parul Chopra, Allie~Del Giorno, Gustavo de~Rosa, Matthew Dixon, Ronen Eldan, Dan Iter, Amit Garg, Abhishek Goswami, Suriya Gunasekar, Emman Haider, Junheng Hao, Russell~J. Hewett, Jamie Huynh, Mojan Javaheripi, Xin Jin, Piero Kauffmann, Nikos Karampatziakis, Dongwoo Kim, Mahoud Khademi, Lev Kurilenko, James~R. Lee, Yin~Tat Lee, Yuanzhi Li, Chen Liang, Weishung Liu, Eric Lin, Zeqi Lin, Piyush Madan, Arindam Mitra, Hardik Modi, Anh Nguyen, Brandon Norick, Barun Patra, Daniel Perez{-}Becker, Thomas Portet, Reid Pryzant, Heyang Qin, Marko Radmilac, Corby Rosset, Sambudha Roy, Olatunji Ruwase, Olli Saarikivi, Amin Saied, Adil Salim, Michael Santacroce, Shital Shah, Ning Shang, Hiteshi Sharma, Xia Song, Masahiro Tanaka,
  Xin Wang, Rachel Ward, Guanhua Wang, Philipp~A. Witte, Michael Wyatt, Can Xu, Jiahang Xu, Sonali Yadav, Fan Yang, Ziyi Yang, Donghan Yu, Chengruidong Zhang, Cyril Zhang, Jianwen Zhang, Li~Lyna Zhang, Yi~Zhang, Yue Zhang, Yunan Zhang, and Xiren Zhou.
\newblock Phi-3 technical report: {A} highly capable language model locally on your phone.
\newblock \emph{CoRR}, abs/2404.14219, 2024.

\bibitem[Bai et~al.(2025)Bai, Chen, Liu, Wang, Ge, Song, Dang, Wang, Wang, Tang, Zhong, Zhu, Yang, Li, Wan, Wang, Ding, Fu, Xu, Ye, Zhang, Xie, Cheng, Zhang, Yang, Xu, and Lin]{Qwen2.5-VL}
Shuai Bai, Keqin Chen, Xuejing Liu, Jialin Wang, Wenbin Ge, Sibo Song, Kai Dang, Peng Wang, Shijie Wang, Jun Tang, Humen Zhong, Yuanzhi Zhu, Ming{-}Hsuan Yang, Zhaohai Li, Jianqiang Wan, Pengfei Wang, Wei Ding, Zheren Fu, Yiheng Xu, Jiabo Ye, Xi~Zhang, Tianbao Xie, Zesen Cheng, Hang Zhang, Zhibo Yang, Haiyang Xu, and Junyang Lin.
\newblock Qwen2.5-vl technical report.
\newblock \emph{CoRR}, abs/2502.13923, 2025.

\bibitem[Beyer et~al.(2024)Beyer, Steiner, Pinto, Kolesnikov, Wang, Salz, Neumann, Alabdulmohsin, Tschannen, Bugliarello, Unterthiner, Keysers, Koppula, Liu, Grycner, Gritsenko, Houlsby, Kumar, Rong, Eisenschlos, Kabra, Bauer, Bosnjak, Chen, Minderer, Voigtlaender, Bica, Balazevic, Puigcerver, Papalampidi, H{\'{e}}naff, Xiong, Soricut, Harmsen, and Zhai]{PaliGemma}
Lucas Beyer, Andreas Steiner, Andr{\'{e}}~Susano Pinto, Alexander Kolesnikov, Xiao Wang, Daniel Salz, Maxim Neumann, Ibrahim Alabdulmohsin, Michael Tschannen, Emanuele Bugliarello, Thomas Unterthiner, Daniel Keysers, Skanda Koppula, Fangyu Liu, Adam Grycner, Alexey~A. Gritsenko, Neil Houlsby, Manoj Kumar, Keran Rong, Julian Eisenschlos, Rishabh Kabra, Matthias Bauer, Matko Bosnjak, Xi~Chen, Matthias Minderer, Paul Voigtlaender, Ioana Bica, Ivana Balazevic, Joan Puigcerver, Pinelopi Papalampidi, Olivier~J. H{\'{e}}naff, Xi~Xiong, Radu Soricut, Jeremiah Harmsen, and Xiaohua Zhai.
\newblock Paligemma: {A} versatile 3b {VLM} for transfer.
\newblock \emph{CoRR}, abs/2407.07726, 2024.

\bibitem[{ByteDance Seed}(2025)]{bytedance2025}
{ByteDance Seed}.
\newblock Seed1.6-embedding.
\newblock Online, 2025.
\newblock URL \url{https://seed1-6-embedding.github.io/}.

\bibitem[Chen et~al.(2025)Chen, Wang, Yang, Zhu, Zhao, Wei, and Dou]{mme5}
Haonan Chen, Liang Wang, Nan Yang, Yutao Zhu, Ziliang Zhao, Furu Wei, and Zhicheng Dou.
\newblock mme5: Improving multimodal multilingual embeddings via high-quality synthetic data.
\newblock In \emph{{ACL} (Findings)}, pp.\  8254--8275. Association for Computational Linguistics, 2025.

\bibitem[Chen et~al.(2024)Chen, Li, Dong, Zhang, He, Wang, Zhao, and Lin]{ShareGPT4V}
Lin Chen, Jinsong Li, Xiaoyi Dong, Pan Zhang, Conghui He, Jiaqi Wang, Feng Zhao, and Dahua Lin.
\newblock Sharegpt4v: Improving large multi-modal models with better captions.
\newblock In \emph{{ECCV} {(17)}}, volume 15075 of \emph{Lecture Notes in Computer Science}, pp.\  370--387. Springer, 2024.

\bibitem[Cherti et~al.(2023)Cherti, Beaumont, Wightman, Wortsman, Ilharco, Gordon, Schuhmann, Schmidt, and Jitsev]{OpenCLIP}
Mehdi Cherti, Romain Beaumont, Ross Wightman, Mitchell Wortsman, Gabriel Ilharco, Cade Gordon, Christoph Schuhmann, Ludwig Schmidt, and Jenia Jitsev.
\newblock Reproducible scaling laws for contrastive language-image learning.
\newblock In \emph{{CVPR}}, pp.\  2818--2829. {IEEE}, 2023.

\bibitem[Faysse et~al.(2025)Faysse, Sibille, Wu, Omrani, Viaud, Hudelot, and Colombo]{ColPali}
Manuel Faysse, Hugues Sibille, Tony Wu, Bilel Omrani, Gautier Viaud, C{\'{e}}line Hudelot, and Pierre Colombo.
\newblock Colpali: Efficient document retrieval with vision language models.
\newblock In \emph{{ICLR}}. OpenReview.net, 2025.

\bibitem[Gao et~al.(2021)Gao, Yao, and Chen]{SimCSE}
Tianyu Gao, Xingcheng Yao, and Danqi Chen.
\newblock Simcse: Simple contrastive learning of sentence embeddings.
\newblock In \emph{{EMNLP} {(1)}}, pp.\  6894--6910. Association for Computational Linguistics, 2021.

\bibitem[Gu et~al.(2024)Gu, Zhang, Zhou, Yu, Xing, Wang, Cao, Jia, Zhang, Wang, Hu, Zhang, Li, Liang, Zhao, Ao, Liu, Feng, and Liu]{Aquila-VL}
Shuhao Gu, Jialing Zhang, Siyuan Zhou, Kevin Yu, Zhaohu Xing, Liangdong Wang, Zhou Cao, Jintao Jia, Zhuoyi Zhang, Yixuan Wang, Zhenchong Hu, Bo{-}Wen Zhang, Jijie Li, Dong Liang, Yingli Zhao, Yulong Ao, Yaoqi Liu, Fangxiang Feng, and Guang Liu.
\newblock Infinity-mm: Scaling multimodal performance with large-scale and high-quality instruction data.
\newblock \emph{CoRR}, abs/2410.18558, 2024.

\bibitem[Gu et~al.(2025{\natexlab{a}})Gu, Yang, Feng, Wang, Zhang, Long, Chen, Cai, and Deng]{UniME}
Tiancheng Gu, Kaicheng Yang, Ziyong Feng, Xingjun Wang, Yanzhao Zhang, Dingkun Long, Yingda Chen, Weidong Cai, and Jiankang Deng.
\newblock Breaking the modality barrier: Universal embedding learning with multimodal llms.
\newblock \emph{CoRR}, abs/2504.17432, 2025{\natexlab{a}}.

\bibitem[Gu et~al.(2025{\natexlab{b}})Gu, Yang, Zhang, An, Feng, Zhang, Cai, Deng, and Bing]{UniME-V2}
Tiancheng Gu, Kaicheng Yang, Kaichen Zhang, Xiang An, Ziyong Feng, Yueyi Zhang, Weidong Cai, Jiankang Deng, and Lidong Bing.
\newblock Unime-v2: Mllm-as-a-judge for universal multimodal embedding learning.
\newblock \emph{CoRR}, abs/2510.13515, 2025{\natexlab{b}}.

\bibitem[Hong et~al.(2024)Hong, Wang, Ding, Yu, Lv, Wang, Cheng, Huang, Ji, Xue, Zhao, Yang, Gu, Zhang, Feng, Yin, Wang, Qi, Song, Zhang, Liu, Xu, Li, Dong, and Tang]{CogVLM2}
Wenyi Hong, Weihan Wang, Ming Ding, Wenmeng Yu, Qingsong Lv, Yan Wang, Yean Cheng, Shiyu Huang, Junhui Ji, Zhao Xue, Lei Zhao, Zhuoyi Yang, Xiaotao Gu, Xiaohan Zhang, Guanyu Feng, Da~Yin, Zihan Wang, Ji~Qi, Xixuan Song, Peng Zhang, Debing Liu, Bin Xu, Juanzi Li, Yuxiao Dong, and Jie Tang.
\newblock Cogvlm2: Visual language models for image and video understanding.
\newblock \emph{CoRR}, abs/2408.16500, 2024.

\bibitem[Hu et~al.(2022)Hu, Shen, Wallis, Allen{-}Zhu, Li, Wang, Wang, and Chen]{LoRA}
Edward~J. Hu, Yelong Shen, Phillip Wallis, Zeyuan Allen{-}Zhu, Yuanzhi Li, Shean Wang, Lu~Wang, and Weizhu Chen.
\newblock Lora: Low-rank adaptation of large language models.
\newblock In \emph{{ICLR}}. OpenReview.net, 2022.

\bibitem[Jiang et~al.(2025)Jiang, Meng, Yang, Yavuz, Zhou, and Chen]{MMEB-v1}
Ziyan Jiang, Rui Meng, Xinyi Yang, Semih Yavuz, Yingbo Zhou, and Wenhu Chen.
\newblock Vlm2vec: Training vision-language models for massive multimodal embedding tasks.
\newblock In \emph{{ICLR}}. OpenReview.net, 2025.

\bibitem[Joshi et~al.(2017)Joshi, Choi, Weld, and Zettlemoyer]{Trivia}
Mandar Joshi, Eunsol Choi, Daniel~S. Weld, and Luke Zettlemoyer.
\newblock Triviaqa: {A} large scale distantly supervised challenge dataset for reading comprehension.
\newblock In \emph{{ACL} {(1)}}, pp.\  1601--1611. Association for Computational Linguistics, 2017.

\bibitem[Ju \& Lee(2025)Ju and Lee]{ju2025generatorembedderharnessinginnate}
Yeong{-}Joon Ju and Seong{-}Whan Lee.
\newblock From generator to embedder: Harnessing innate abilities of multimodal llms via building zero-shot discriminative embedding model.
\newblock \emph{CoRR}, abs/2508.00955, 2025.

\bibitem[Kong et~al.(2025)Kong, Zhang, Liu, Zhang, Feng, Yang, Wang, Tian, W., Zhang, and Zhou]{UNITE}
Fanheng Kong, Jingyuan Zhang, Yahui Liu, Hongzhi Zhang, Shi Feng, Xiaocui Yang, Daling Wang, Yu~Tian, Victoria W., Fuzheng Zhang, and Guorui Zhou.
\newblock Modality curation: Building universal embeddings for advanced multimodal information retrieval.
\newblock \emph{CoRR}, abs/2505.19650, 2025.

\bibitem[Kwiatkowski et~al.(2019)Kwiatkowski, Palomaki, Redfield, Collins, Parikh, Alberti, Epstein, Polosukhin, Devlin, Lee, Toutanova, Jones, Kelcey, Chang, Dai, Uszkoreit, Le, and Petrov]{NQ}
Tom Kwiatkowski, Jennimaria Palomaki, Olivia Redfield, Michael Collins, Ankur~P. Parikh, Chris Alberti, Danielle Epstein, Illia Polosukhin, Jacob Devlin, Kenton Lee, Kristina Toutanova, Llion Jones, Matthew Kelcey, Ming{-}Wei Chang, Andrew~M. Dai, Jakob Uszkoreit, Quoc Le, and Slav Petrov.
\newblock Natural questions: a benchmark for question answering research.
\newblock \emph{Trans. Assoc. Comput. Linguistics}, 7:\penalty0 452--466, 2019.

\bibitem[Lan et~al.(2025)Lan, Niu, Meng, Zhou, and Su]{LLaVE}
Zhibin Lan, Liqiang Niu, Fandong Meng, Jie Zhou, and Jinsong Su.
\newblock Llave: Large language and vision embedding models with hardness-weighted contrastive learning.
\newblock \emph{CoRR}, abs/2503.04812, 2025.

\bibitem[Li et~al.(2025)Li, Zhang, Guo, Zhang, Li, Zhang, Zhang, Zhang, Li, Liu, and Li]{LLaVA-OneVision}
Bo~Li, Yuanhan Zhang, Dong Guo, Renrui Zhang, Feng Li, Hao Zhang, Kaichen Zhang, Peiyuan Zhang, Yanwei Li, Ziwei Liu, and Chunyuan Li.
\newblock Llava-onevision: Easy visual task transfer.
\newblock \emph{Trans. Mach. Learn. Res.}, 2025, 2025.

\bibitem[Li et~al.(2022)Li, Li, Xiong, and Hoi]{BLIP}
Junnan Li, Dongxu Li, Caiming Xiong, and Steven C.~H. Hoi.
\newblock {BLIP:} bootstrapping language-image pre-training for unified vision-language understanding and generation.
\newblock In \emph{{ICML}}, volume 162 of \emph{Proceedings of Machine Learning Research}, pp.\  12888--12900. {PMLR}, 2022.

\bibitem[Li et~al.(2023{\natexlab{a}})Li, Li, Savarese, and Hoi]{BLIP-2}
Junnan Li, Dongxu Li, Silvio Savarese, and Steven C.~H. Hoi.
\newblock {BLIP-2:} bootstrapping language-image pre-training with frozen image encoders and large language models.
\newblock In \emph{{ICML}}, volume 202 of \emph{Proceedings of Machine Learning Research}, pp.\  19730--19742. {PMLR}, 2023{\natexlab{a}}.

\bibitem[Li et~al.(2023{\natexlab{b}})Li, Zhang, Zhang, Long, Xie, and Zhang]{GTE}
Zehan Li, Xin Zhang, Yanzhao Zhang, Dingkun Long, Pengjun Xie, and Meishan Zhang.
\newblock Towards general text embeddings with multi-stage contrastive learning.
\newblock \emph{CoRR}, abs/2308.03281, 2023{\natexlab{b}}.

\bibitem[Li et~al.(2023{\natexlab{c}})Li, Li, Yang, Zhao, Song, Luo, Li, and Yang]{li2023curriculum}
Zheng Li, Xiang Li, Lingfeng Yang, Borui Zhao, Renjie Song, Lei Luo, Jun Li, and Jian Yang.
\newblock Curriculum temperature for knowledge distillation.
\newblock In \emph{Proceedings of the AAAI Conference on Artificial Intelligence}, volume~37, pp.\  1504--1512, 2023{\natexlab{c}}.

\bibitem[Lin et~al.(2025)Lin, Lee, Shoeybi, Lin, Catanzaro, and Ping]{Mm-Embed}
Sheng{-}Chieh Lin, Chankyu Lee, Mohammad Shoeybi, Jimmy Lin, Bryan Catanzaro, and Wei Ping.
\newblock Mm-embed: Universal multimodal retrieval with multimodal {LLMS}.
\newblock In \emph{{ICLR}}. OpenReview.net, 2025.

\bibitem[Liu et~al.(2023)Liu, Li, Wu, and Lee]{LLaVA}
Haotian Liu, Chunyuan Li, Qingyang Wu, and Yong~Jae Lee.
\newblock Visual instruction tuning.
\newblock In \emph{NeurIPS}, 2023.

\bibitem[Liu et~al.(2024)Liu, Li, Li, Li, Zhang, Shen, and Lee]{LLaVA-NeXT}
Haotian Liu, Chunyuan Li, Yuheng Li, Bo~Li, Yuanhan Zhang, Sheng Shen, and Yong~Jae Lee.
\newblock Llava-next: Improved reasoning, ocr, and world knowledge, January 2024.
\newblock URL \url{https://llava-vl.github.io/blog/2024-01-30-llava-next/}.

\bibitem[Meng et~al.(2025)Meng, Jiang, Liu, Su, Yang, Fu, Qin, Chen, Xu, Xiong, Zhou, Chen, and Yavuz]{MMEB-v2}
Rui Meng, Ziyan Jiang, Ye~Liu, Mingyi Su, Xinyi Yang, Yuepeng Fu, Can Qin, Zeyuan Chen, Ran Xu, Caiming Xiong, Yingbo Zhou, Wenhu Chen, and Semih Yavuz.
\newblock Vlm2vec-v2: Advancing multimodal embedding for videos, images, and visual documents.
\newblock \emph{CoRR}, abs/2507.04590, 2025.

\bibitem[Nguyen et~al.(2016)Nguyen, Rosenberg, Song, Gao, Tiwary, Majumder, and Deng]{MARCO}
Tri Nguyen, Mir Rosenberg, Xia Song, Jianfeng Gao, Saurabh Tiwary, Rangan Majumder, and Li~Deng.
\newblock {MS} {MARCO:} {A} human generated machine reading comprehension dataset.
\newblock In \emph{CoCo@NIPS}, volume 1773 of \emph{{CEUR} Workshop Proceedings}. CEUR-WS.org, 2016.

\bibitem[Qiu et~al.(2023)Qiu, Hu, Yuan, Zhou, Zhang, and Yang]{qiu2023semanticscreatedequalcontrastive}
Zi{-}Hao Qiu, Quanqi Hu, Zhuoning Yuan, Denny Zhou, Lijun Zhang, and Tianbao Yang.
\newblock Not all semantics are created equal: Contrastive self-supervised learning with automatic temperature individualization.
\newblock In \emph{{ICML}}, volume 202 of \emph{Proceedings of Machine Learning Research}, pp.\  28389--28421. {PMLR}, 2023.

\bibitem[Radford et~al.(2021)Radford, Kim, Hallacy, Ramesh, Goh, Agarwal, Sastry, Askell, Mishkin, Clark, Krueger, and Sutskever]{CLIP}
Alec Radford, Jong~Wook Kim, Chris Hallacy, Aditya Ramesh, Gabriel Goh, Sandhini Agarwal, Girish Sastry, Amanda Askell, Pamela Mishkin, Jack Clark, Gretchen Krueger, and Ilya Sutskever.
\newblock Learning transferable visual models from natural language supervision.
\newblock In \emph{{ICML}}, volume 139 of \emph{Proceedings of Machine Learning Research}, pp.\  8748--8763. {PMLR}, 2021.

\bibitem[Rajpurkar et~al.(2016)Rajpurkar, Zhang, Lopyrev, and Liang]{SQuAD}
Pranav Rajpurkar, Jian Zhang, Konstantin Lopyrev, and Percy Liang.
\newblock Squad: 100, 000+ questions for machine comprehension of text.
\newblock In \emph{{EMNLP}}, pp.\  2383--2392. The Association for Computational Linguistics, 2016.

\bibitem[Robinson et~al.(2021)Robinson, Chuang, Sra, and Jegelka]{robinson2021contrastivelearninghardnegative}
Joshua~David Robinson, Ching{-}Yao Chuang, Suvrit Sra, and Stefanie Jegelka.
\newblock Contrastive learning with hard negative samples.
\newblock In \emph{{ICLR}}. OpenReview.net, 2021.

\bibitem[Rusak et~al.(2025)Rusak, Reizinger, Juhos, Bringmann, Zimmermann, and Brendel]{rusak2025infonceidentifyinggaptheory}
Evgenia Rusak, Patrik Reizinger, Attila Juhos, Oliver Bringmann, Roland~S. Zimmermann, and Wieland Brendel.
\newblock Infonce: Identifying the gap between theory and practice.
\newblock In \emph{{AISTATS}}, volume 258 of \emph{Proceedings of Machine Learning Research}, pp.\  4159--4167. {PMLR}, 2025.

\bibitem[Schuhmann et~al.(2022)Schuhmann, Beaumont, Vencu, Gordon, Wightman, Cherti, Coombes, Katta, Mullis, Wortsman, Schramowski, Kundurthy, Crowson, Schmidt, Kaczmarczyk, and Jitsev]{LAION}
Christoph Schuhmann, Romain Beaumont, Richard Vencu, Cade Gordon, Ross Wightman, Mehdi Cherti, Theo Coombes, Aarush Katta, Clayton Mullis, Mitchell Wortsman, Patrick Schramowski, Srivatsa Kundurthy, Katherine Crowson, Ludwig Schmidt, Robert Kaczmarczyk, and Jenia Jitsev.
\newblock {LAION-5B:} an open large-scale dataset for training next generation image-text models.
\newblock In \emph{NeurIPS}, 2022.

\bibitem[Thirukovalluru et~al.(2025)Thirukovalluru, Meng, Liu, K, Su, Nie, Yavuz, Zhou, Chen, and Dhingra]{B3}
Raghuveer Thirukovalluru, Rui Meng, Ye~Liu, Karthikeyan K, Mingyi Su, Ping Nie, Semih Yavuz, Yingbo Zhou, Wenhu Chen, and Bhuwan Dhingra.
\newblock Breaking the batch barrier {(B3)} of contrastive learning via smart batch mining.
\newblock \emph{CoRR}, abs/2505.11293, 2025.

\bibitem[Thorne et~al.(2018)Thorne, Vlachos, Christodoulopoulos, and Mittal]{FEVER}
James Thorne, Andreas Vlachos, Christos Christodoulopoulos, and Arpit Mittal.
\newblock {FEVER:} a large-scale dataset for fact extraction and verification.
\newblock In \emph{{NAACL-HLT}}, pp.\  809--819. Association for Computational Linguistics, 2018.

\bibitem[Vera et~al.(2025)Vera, Dua, Zhang, Salz, Mullins, Panyam, Smoot, Naim, Zou, Chen, Cer, Lisak, Choi, Gonzalez, Sanseviero, Cameron, Ballantyne, Black, Chen, Wang, Li, Martins, Lee, Sherwood, Ji, Wu, Zheng, Singh, Sharma, Sreepathihalli, Jain, Elarabawy, Co, Doumanoglou, Samari, Hora, Potetz, Kim, Alfonseca, Moiseev, Han, Gomez, {\'{A}}brego, Zhang, Hui, Han, Gill, Chen, Chen, Shanbhogue, Boratko, Suganthan, Duddu, Mariserla, Ariafar, Zhang, Zhang, Baumgartner, Goenka, Qiu, Dabral, Walker, Rao, Khawaja, Zhou, Ren, Xia, Chen, Chen, Dong, Ding, Visin, Liu, Zhang, Kenealy, Casbon, Kumar, Mesnard, Gleicher, Brick, Lacombe, Roberts, Yin, Sung, Hoffmann, Warkentin, Joulin, Duerig, and Seyedhosseini]{EmbeddingGemma}
Henrique~Schechter Vera, Sahil Dua, Biao Zhang, Daniel Salz, Ryan Mullins, Sindhu~Raghuram Panyam, Sara Smoot, Iftekhar Naim, Joe Zou, Feiyang Chen, Daniel Cer, Alice Lisak, Min Choi, Lucas Gonzalez, Omar Sanseviero, Glenn Cameron, Ian Ballantyne, Kat Black, Kaifeng Chen, Weiyi Wang, Zhe Li, Gus Martins, Jinhyuk Lee, Mark Sherwood, Ju{-}yeong Ji, Renjie Wu, Jingxiao Zheng, Jyotinder Singh, Abheesht Sharma, Divyashree Sreepathihalli, Aashi Jain, Adham Elarabawy, AJ~Co, Andreas Doumanoglou, Babak Samari, Ben Hora, Brian Potetz, Dahun Kim, Enrique Alfonseca, Fedor Moiseev, Feng Han, Frank~Palma Gomez, Gustavo~Hern{\'{a}}ndez {\'{A}}brego, Hesen Zhang, Hui Hui, Jay Han, Karan Gill, Ke~Chen, Koert Chen, Madhuri Shanbhogue, Michael Boratko, Paul Suganthan, Sai Meher~Karthik Duddu, Sandeep Mariserla, Setareh Ariafar, Shanfeng Zhang, Shijie Zhang, Simon Baumgartner, Sonam Goenka, Steve Qiu, Tanmaya Dabral, Trevor Walker, Vikram Rao, Waleed Khawaja, Wenlei Zhou, Xiaoqi Ren, Ye~Xia, Yichang Chen, Yi{-}Ting Chen, Zhe
  Dong, Zhongli Ding, Francesco Visin, Ga{\"{e}}l Liu, Jiageng Zhang, Kathleen Kenealy, Michelle Casbon, Ravin Kumar, Thomas Mesnard, Zach Gleicher, Cormac Brick, Olivier Lacombe, Adam Roberts, Qin Yin, Yun{-}Hsuan Sung, Raphael Hoffmann, Tris Warkentin, Armand Joulin, Tom Duerig, and Mojtaba Seyedhosseini.
\newblock Embeddinggemma: Powerful and lightweight text representations.
\newblock \emph{CoRR}, abs/2509.20354, 2025.

\bibitem[Wang et~al.(2024{\natexlab{a}})Wang, Yang, Huang, Yang, Majumder, and Wei]{GME_ref_63}
Liang Wang, Nan Yang, Xiaolong Huang, Linjun Yang, Rangan Majumder, and Furu Wei.
\newblock Improving text embeddings with large language models.
\newblock In \emph{{ACL} {(1)}}, pp.\  11897--11916. Association for Computational Linguistics, 2024{\natexlab{a}}.

\bibitem[Wang et~al.(2024{\natexlab{b}})Wang, Bai, Tan, Wang, Fan, Bai, Chen, Liu, Wang, Ge, Fan, Dang, Du, Ren, Men, Liu, Zhou, Zhou, and Lin]{Qwen2-VL}
Peng Wang, Shuai Bai, Sinan Tan, Shijie Wang, Zhihao Fan, Jinze Bai, Keqin Chen, Xuejing Liu, Jialin Wang, Wenbin Ge, Yang Fan, Kai Dang, Mengfei Du, Xuancheng Ren, Rui Men, Dayiheng Liu, Chang Zhou, Jingren Zhou, and Junyang Lin.
\newblock Qwen2-vl: Enhancing vision-language model's perception of the world at any resolution.
\newblock \emph{CoRR}, abs/2409.12191, 2024{\natexlab{b}}.

\bibitem[Wang et~al.(2025)Wang, Gao, Gu, Pu, Cui, Wei, Liu, Jing, Ye, Shao, Wang, Chen, Zhang, Yang, Wang, Wei, Yin, Li, Cui, Chen, Ding, Tian, Wu, Xie, Li, Yang, Duan, Wang, Hou, Hao, Zhang, Li, Zhao, Duan, Deng, Fu, He, Wang, He, Shi, He, Xiong, Lv, Wu, Shao, Zhang, Deng, Qi, Ge, Guo, Zhang, Zhang, Cao, Lin, Tang, Gao, Huang, Gu, Lyu, Tang, Wang, Lv, Ouyang, Wang, Dou, Zhu, Lu, Lin, Dai, Su, Zhou, Chen, Qiao, Wang, and Luo]{InternVL3.5}
Weiyun Wang, Zhangwei Gao, Lixin Gu, Hengjun Pu, Long Cui, Xingguang Wei, Zhaoyang Liu, Linglin Jing, Shenglong Ye, Jie Shao, Zhaokai Wang, Zhe Chen, Hongjie Zhang, Ganlin Yang, Haomin Wang, Qi~Wei, Jinhui Yin, Wenhao Li, Erfei Cui, Guanzhou Chen, Zichen Ding, Changyao Tian, Zhenyu Wu, JingJing Xie, Zehao Li, Bowen Yang, Yuchen Duan, Xuehui Wang, Zhi Hou, Haoran Hao, Tianyi Zhang, Songze Li, Xiangyu Zhao, Haodong Duan, Nianchen Deng, Bin Fu, Yinan He, Yi~Wang, Conghui He, Botian Shi, Junjun He, Yingtong Xiong, Han Lv, Lijun Wu, Wenqi Shao, Kaipeng Zhang, Huipeng Deng, Biqing Qi, Jiaye Ge, Qipeng Guo, Wenwei Zhang, Songyang Zhang, Maosong Cao, Junyao Lin, Kexian Tang, Jianfei Gao, Haian Huang, Yuzhe Gu, Chengqi Lyu, Huanze Tang, Rui Wang, Haijun Lv, Wanli Ouyang, Limin Wang, Min Dou, Xizhou Zhu, Tong Lu, Dahua Lin, Jifeng Dai, Weijie Su, Bowen Zhou, Kai Chen, Yu~Qiao, Wenhai Wang, and Gen Luo.
\newblock Internvl3.5: Advancing open-source multimodal models in versatility, reasoning, and efficiency.
\newblock \emph{CoRR}, abs/2508.18265, 2025.

\bibitem[Wei et~al.(2024)Wei, Chen, Chen, Hu, Zhang, Fu, Ritter, and Chen]{UniIR}
Cong Wei, Yang Chen, Haonan Chen, Hexiang Hu, Ge~Zhang, Jie Fu, Alan Ritter, and Wenhu Chen.
\newblock Uniir: Training and benchmarking universal multimodal information retrievers.
\newblock In \emph{{ECCV} {(87)}}, volume 15145 of \emph{Lecture Notes in Computer Science}, pp.\  387--404. Springer, 2024.

\bibitem[Xiao et~al.(2024)Xiao, Wu, Xu, Dai, Hu, Lu, Zeng, Liu, and Yuan]{Florence-2}
Bin Xiao, Haiping Wu, Weijian Xu, Xiyang Dai, Houdong Hu, Yumao Lu, Michael Zeng, Ce~Liu, and Lu~Yuan.
\newblock Florence-2: Advancing a unified representation for a variety of vision tasks.
\newblock In \emph{{CVPR}}, pp.\  4818--4829. {IEEE}, 2024.

\bibitem[Xie et~al.(2025)Xie, Wang, Kong, Li, Liang, Zhang, Leng, and Yin]{xie2025fg}
Chunyu Xie, Bin Wang, Fanjing Kong, Jincheng Li, Dawei Liang, Gengshen Zhang, Dawei Leng, and Yuhui Yin.
\newblock Fg-clip: Fine-grained visual and textual alignment.
\newblock \emph{arXiv preprint arXiv:2505.05071}, 2025.

\bibitem[Xue et~al.(2025)Xue, Li, and Liu]{QQMM-embed}
Youze Xue, Dian Li, and Gang Liu.
\newblock Improve multi-modal embedding learning via explicit hard negative gradient amplifying.
\newblock \emph{CoRR}, abs/2506.02020, 2025.

\bibitem[Yang et~al.(2018)Yang, Qi, Zhang, Bengio, Cohen, Salakhutdinov, and Manning]{HotpotQA}
Zhilin Yang, Peng Qi, Saizheng Zhang, Yoshua Bengio, William~W. Cohen, Ruslan Salakhutdinov, and Christopher~D. Manning.
\newblock Hotpotqa: {A} dataset for diverse, explainable multi-hop question answering.
\newblock In \emph{{EMNLP}}, pp.\  2369--2380. Association for Computational Linguistics, 2018.

\bibitem[Yu et~al.(2025)Yu, Zhao, Yan, Korycki, Wang, He, Liu, Zhang, Fan, and Yu]{CAFe}
Hao Yu, Zhuokai Zhao, Shen Yan, Lukasz Korycki, Jianyu Wang, Baosheng He, Jiayi Liu, Lizhu Zhang, Xiangjun Fan, and Hanchao Yu.
\newblock Cafe: Unifying representation and generation with contrastive-autoregressive finetuning.
\newblock \emph{CoRR}, abs/2503.19900, 2025.

\bibitem[Zhai et~al.(2023)Zhai, Mustafa, Kolesnikov, and Beyer]{SigLIP}
Xiaohua Zhai, Basil Mustafa, Alexander Kolesnikov, and Lucas Beyer.
\newblock Sigmoid loss for language image pre-training.
\newblock In \emph{{ICCV}}, pp.\  11941--11952. {IEEE}, 2023.

\bibitem[Zhang et~al.(2024)Zhang, Luan, Hu, Lee, Qiao, Chen, Su, and Chang]{MagicLens}
Kai Zhang, Yi~Luan, Hexiang Hu, Kenton Lee, Siyuan Qiao, Wenhu Chen, Yu~Su, and Ming{-}Wei Chang.
\newblock Magiclens: Self-supervised image retrieval with open-ended instructions.
\newblock In \emph{{ICML}}. OpenReview.net, 2024.

\bibitem[Zhang et~al.(2025)Zhang, Zhang, Xie, Li, Dai, Long, Xie, Zhang, Li, and Zhang]{GME}
Xin Zhang, Yanzhao Zhang, Wen Xie, Mingxin Li, Ziqi Dai, Dingkun Long, Pengjun Xie, Meishan Zhang, Wenjie Li, and Min Zhang.
\newblock Bridging modalities: Improving universal multimodal retrieval by multimodal large language models.
\newblock In \emph{{CVPR}}, pp.\  9274--9285. Computer Vision Foundation / {IEEE}, 2025.

\bibitem[Zhou et~al.(2025)Zhou, Xiong, Liu, Liu, Xiao, Wang, Zhao, Zhang, and Lian]{MegaPairs}
Junjie Zhou, Yongping Xiong, Zheng Liu, Ze~Liu, Shitao Xiao, Yueze Wang, Bo~Zhao, Chen~Jason Zhang, and Defu Lian.
\newblock Megapairs: Massive data synthesis for universal multimodal retrieval.
\newblock In \emph{{ACL} {(1)}}, pp.\  19076--19095. Association for Computational Linguistics, 2025.

\end{thebibliography}
\bibliographystyle{colm2025_conference}
%

\end{document}